\pdfoutput=1
\documentclass[11pt]{article}
\usepackage{coling2020}
\usepackage{times}
\usepackage{url}
\usepackage{latexsym}

\usepackage{bm}
\usepackage{amsmath}
\usepackage{amsfonts}
\usepackage{booktabs}

\usepackage{array}
\usepackage{color}
\usepackage{graphicx}
\usepackage{epstopdf}
\usepackage{multirow}
\usepackage{microtype}
\usepackage{float}

\colingfinalcopy 

\title{Exploiting Rich Syntax for Better Knowledge Base Question Answering}

\author{
	Pengju Zhang$^1$, Yonghui Jia$^1$, Muhua Zhu$^2$, WenLiang Chen$^1$, Min Zhang$^1$ \\  \\
	$^1$Institute of Artificial Intelligence, School of Computer Science and Technology, \\ Soochow University, China\\
	$^2$Tencent, China \\
	{\tt pjzhang\_stu@qq.com}\\
	{\tt yhjia2018@163.com} \\
	{\tt zhumuhua@gmail.com}\\
	{\tt \{wlchen, minzhang\}@suda.edu.cn}\\
}

\date{}

\begin{document}
\maketitle
\begin{abstract}

Recent studies on Knowledge Base Question Answering (KBQA) have shown great progress on this task via better question understanding.
Previous works for encoding questions mainly focus on the word sequences, but seldom consider the information from syntactic trees.
In this paper, we propose an approach to learn syntax-based representations for KBQA. 
First, we encode path-based syntax by considering the shortest dependency paths between keywords. Then, we propose two encoding strategies 
to mode the information of whole syntactic trees to obtain tree-based syntax. Finally, we combine both path-based and tree-based syntax representations for KBQA.
We conduct extensive experiments on a widely used benchmark
dataset and the experimental results show that our syntax-aware systems can
make full use of syntax information in different settings and achieve state-of-the-art performance of KBQA.
    
\end{abstract}


\section{Introduction}
Question answering over knowledge base (KBQA) is an appealing task that arises with the advent of large-scale knowledge bases such as Freebase~\cite{bollackeretal:2008},  DBpedia~\cite{auer2007dbpedia}, and YAGO~\cite{suchaneketal:2007}. Briefly, this task is defined to be taking a natural language question as input and searching for an entity or an attribute value as an answer returned from knowledge bases. For example, the question ``{\em what movies did Diana play in?}" can be answered by translating the question into a triplet query {\em (Diana, play, ?)} that will be issued to a knowledge base. One key component of KBQA is question understanding which aims to transform a question into some semantic representation.

Approaches to question understanding can be broadly categorized into two groups. One group relies on the use of semantic parsing  techniques~\cite{Berant2013SemanticPO,reddy-etal-2014-large}. The main idea is to transform a natural language question into a formal meaning representation, eg., lambda calculus. The other group of approaches instead focus on generating query graphs directly with the aid of knowledge bases~\cite{Yih2015SemanticPV,Bao2016ConstraintBasedQA,Hu2018AnsweringNL,Luo2018KnowledgeBQ}.
Both solutions have gained a lot of attention, which are capable of obtaining plenty of essential question information.



\begin{figure*}
  \centering
  \includegraphics[width=12cm]{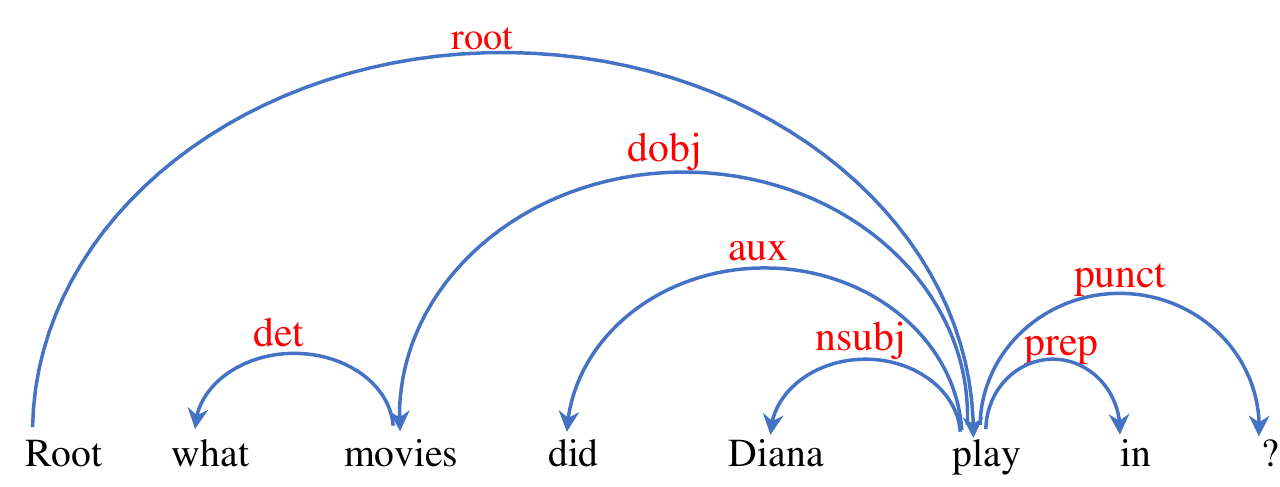}
  \caption{An example of dependency tree provided by the Stanford Parser.}
  \label{fig:dep_res}
\end{figure*}

In recent years, the syntax information has also been investigated for better question understanding in KBQA since it is able to capture the connections between long distance words. One
representative approach is to use the shortest dependency path (SDP) from the answer word to an entity~\cite{Luo2018KnowledgeBQ}. This approach has brought remarkable results, since the path words are indeed important for understanding the questions.
However, it mainly concerns the
connections between words, while neglects the representation of whole syntactic trees.  There exist tree-based syntax representations and it is interesting to examine whether incorporating richer syntactic information can further improve system performance. In this paper, we follow this direction and suggest to utilize dependency syntax to enhance the representation of questions in KBQA task. Dependency syntax is a widely used syntactic parsing formalism that captures interdependence between the words in a sentence, as illustrated in Figure~\ref{fig:dep_res}.

In order to comprehensively explore the impact of syntactic information on KBQA performance, we propose to utilize three different methods to capture syntactic information, which are shortest dependency path (SDP), tree position feature (TPF), and Tree-GRU feature, respectively. The methods are designed from two different perspectives which consider path-based information and tree-based information respectively. Extensive experiments on a benchmark dataset show that the syntactic representations achieve significant improvements over the baseline system. By combining the syntax-based representations, we can further improve the performance of our systems. The contributions of the paper are as follows.

\begin{itemize}

\item We experiment with two types of syntax-based representations, path-based and tree-based syntax representations to better encode the information of questions for KBQA. There are two methods
for modeling whole syntax trees to obtain tree-based representations. To the best of our knowledge, it is the first time that the information from entire syntax trees is used for KBQA.

\item We utilize the syntax-based representations for KBQA in different settings. The experimental results show that the proposed approach can effectively improve the performance of a state-of-the-art baseline system.

\end{itemize}

\section{The Baseline System}
We use the system in Luo et al.~\shortcite{Luo2018KnowledgeBQ} as our baseline system where the syntactic feature of shortest dependency path is omitted.  We describe the system briefly here and for more  details please refer to ~\newcite{Luo2018KnowledgeBQ}.

\subsection{Query Graph Generation}
\begin{figure*}
    \centering
    \includegraphics[width=16cm]{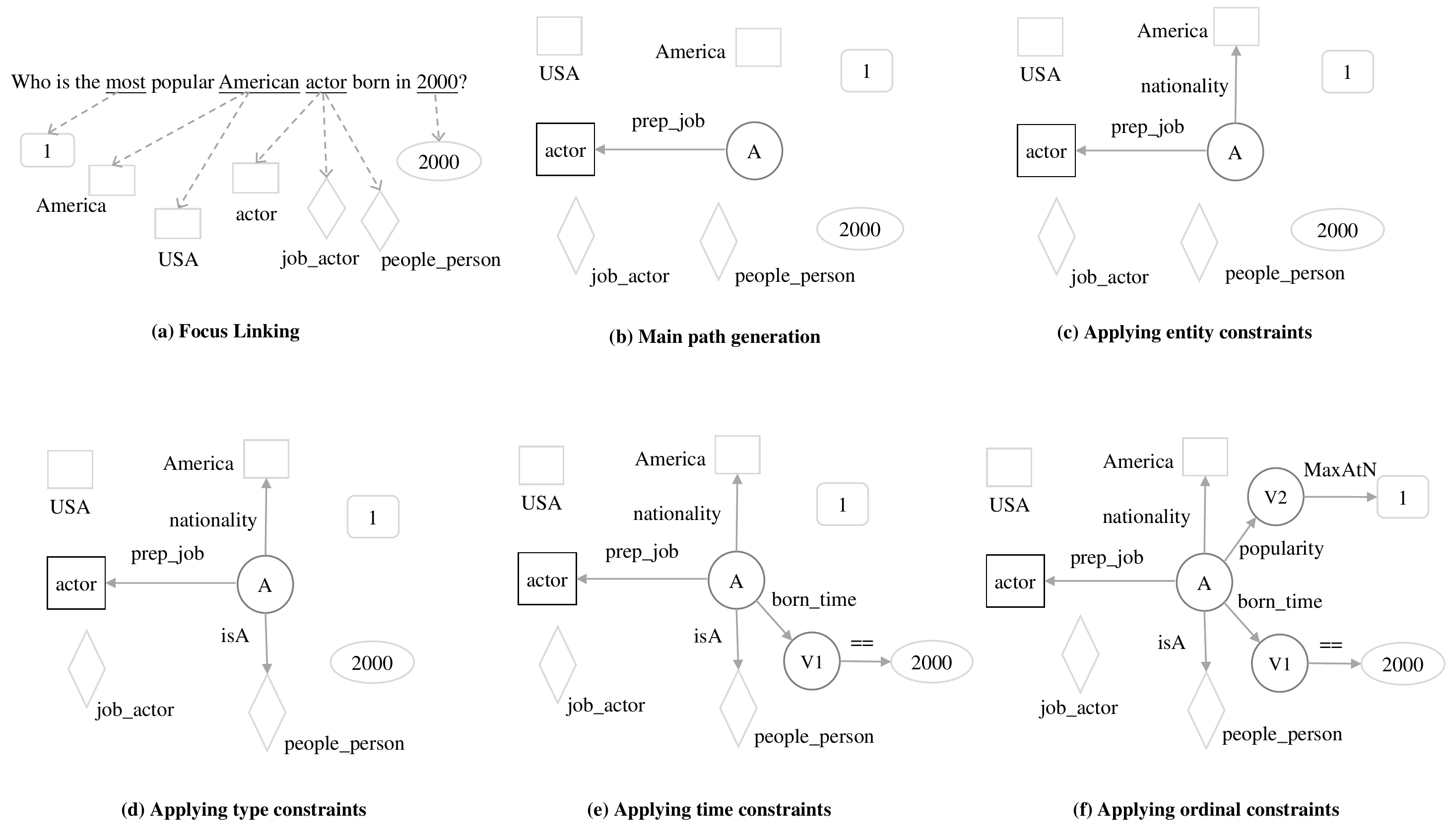}
    \caption{Query graph generation process of `` Who is the most popular American actor born in 2000? ".}
    \label{fig:graph_generation}
\end{figure*}
A query graph is a representation of question semantics that is used to query a knowledge base for an answer. Given a question, we first carry out focus linking to obtain four types of semantic constraints, including \textbf{entity}, implicit \textbf{type}, \textbf{time} interval, and \textbf{ordinal}. For entity linking, we resort to the tool SMART~\cite{smart:2015} to obtain (mention, entity) pairs. 
For type linking, we use word embeddings to calculate the similarity between consecutive sub-sequences in the question (up to three words) and all the type words in the knowledge base, and select the top-10 (mention, type) pairs according to similarity scores. Regarding time word linking, we use regular expression matching to extract time information. Regarding ordinal number linking, we use a predefined ordinal vocabulary and  ``ordinal number + superlative" pattern to extract the integer expressions.~\footnote{About 20 superlative words, such as largest, highest, latest.}  
An example result of focus linking is shown in Figure~\ref{fig:graph_generation}(a).
	
After focus linking, we perform a one-hop and two-hop search based on the linked entity words to get the main paths, as shown in Figure~\ref{fig:graph_generation}(b). Next, we add the unused words as one of the four types of constrains onto the nodes in the main path. For entity constraints, we use depth-first search method to add possible entity constraints as one-hop predicate to the main path, as exemplified in Figure~\ref{fig:graph_generation}(c). For type constraints, we simply add type words to the answer node (Figure~\ref{fig:graph_generation}(d)). For time constraints, we add the time-interval words to the main path where constraints are in the form of two hops and the predicate in the second hop is required to be a virtual predicate such as comparison operators ``=='', ``\textgreater" and ``\textless" (Figure~\ref{fig:graph_generation}(e)). For ordinal constraints, we add ordinal numbers to the main path where constraints are also two-hop and the predicate of the second hop indicates increasing or descending order (Figure~\ref{fig:graph_generation}(f)). 

\subsection{Semantic Matching}
 In the matching process, we encode the question and the query graphs separately, and calculate the similarity between the question and query graphs.
	
\subsubsection{Question Encoding}\label{sec:basicencode}
We first anonymize entity mentions and time words in the question with the dummy symbol ``$<$E$>$" and ``$<$Tm$>$" respectively. When encoding the question, we use a pre-trained embedding matrix $\bm{E}_w\in \mathbb{R}^{|\bm{V}_w | \times d}$ to initialize each word in the question, where $|\bm{V}_w|$ is the size of vocabulary and $d$ is the dimension size of vectors. Then we apply Bidirectional GRU (BiGRU)~\cite{choetal:2014} to encode words of the question, and concatenate the last hidden states of the forward and backward process of BiGRU as the representation of the whole question, denoted as $\bm{q}= [\overset{\rightarrow}{\bm{h}_q},\overset{\leftarrow}{\bm{h}_q}]$.

\subsubsection{Query Graph Encoding}


For a query graph, we split it into sub-paths starting from the answer node to each focus node. For each sub-path, we only encode the relation on it. The relation is usually a combined word string $sp^{(id)}$, such as ``located\_in". 
Each sub-path is encoded from two aspects. (1) We treat the combined word string as one token. Then, we use the vector matrix $\bm{E}_{sp} \in \mathbb{R}^{|\bm{V}_{sp} | \times d}$ to convert it into the corresponding vector expression ${\bm{sp}^{(id)}}$, where $\bm{E}_{sp}$ is randomly initialized and $|\bm{V}_{sp}|$ is the size of the relations table. (2) We split the combined word string into normal words, such as ``located" and ``in". For normal words $\{{sp}^{(w_1)},…,{sp}^{(w_n)}\}$, we use the pre-trained embedding matrix $\bm{E}_w$ to convert them into corresponding vector $\{\bm{sp}^{(w_1)},…,\bm{sp}^{(w_n)}\}$. Then we use the average vector to represent the normal words $\bm{sp}^{(w)}=\frac{1}{n} \sum_{i=1}^{n}\bm{sp}^{(w_i)}$. Therefore, we accumulate $\bm{sp}^{(w)}$ and $\bm{sp}^{(id)}$ to obtain the representation of each sub-path $\bm{sp}$. 
Finally, we use the maxpooling strategy to get the representation of the query graph $\bm{p}$ based on sub-paths.

\subsubsection{Similarity Calculation}
We take the cosine similarity between the question $q$ and the query graph $p$ encoding as a semantic score $S_{rm}(q,p)$ which is a feature to calculate the final similarity $S(Q,G)$. To compute the similarity $S(Q,G)$ between a question $Q$ and a query graph $G$, we weight and sum a feature based on semantic score ($S_{rm}(q,p)$) and other features as described in~\newcite{Luo2018KnowledgeBQ}.

\subsection{Training} 
Then we use Hinge Loss as the objective function to maximize the margin between positive query graph $G^+$ and negative query graph $G^-$:
\begin{equation}
\text {loss}=\max \{0, \lambda-S(Q, G^{+})+S(Q, G^{-})\},
\end{equation}
where $\lambda$ is the parameter of the margin which we set to 0.5. We decide whether a query graph is positive or negative according to the F1 score that the query graph can achieve. That is, if the F1 score of a query graph is greater than a threshold, then it is regarded as positive $G^+$, otherwise as a negative example $G^-$.


\begin{figure*}
  \centering
  \includegraphics[width=16cm]{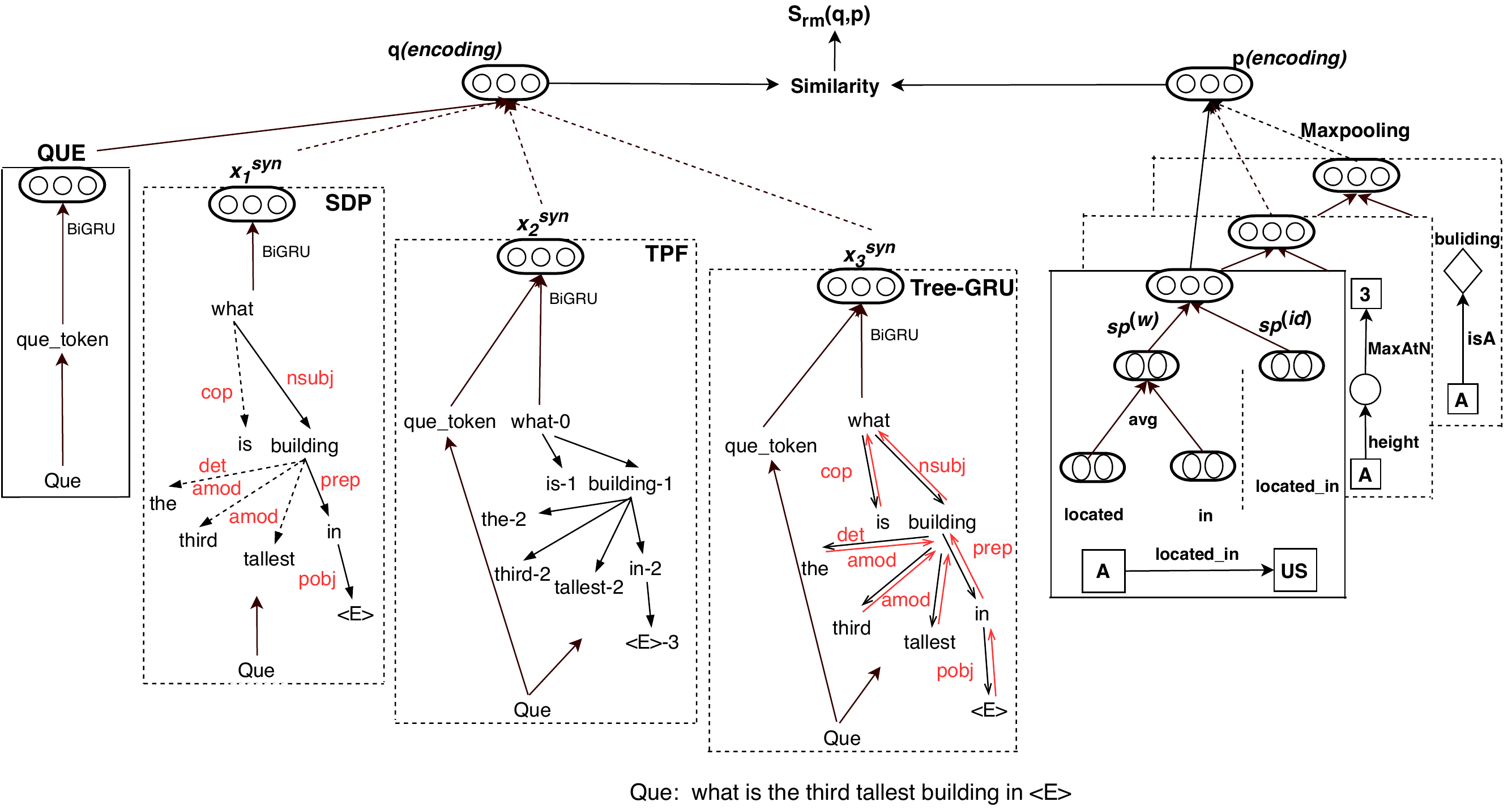}
  \caption{The main structure diagram of our KBQA system. The methods and conditions in the dotted frame are not necessarily used.}
  \label{fig:querygraph}
\end{figure*}

\section{Syntax-Aware Approach}
In this section, we describe three different  methods to encode syntactic information of a question from diverse perspectives. 
\begin{itemize}
  \item SDP (shortest dependency path) is the shortest path between the target word $w_{i}$ and the word $w_{a}$ that refers to the answer node in the parse tree. See details in Section~\ref{sdp_method}.
 \item TPF (tree position feature): For each word $w_{i}$ in a question, we compute the distance of the word to the root word $r_i$ in the parse tree. Such distance is regarded as a position feature. See details in Section~\ref{tpf_method}.
  \item Tree-GRU (tree-structured gated recurrent unit): After obtaining the dependency tree of a question, we encode the whole tree by using Tree-GRU and obtain a vector representation for each word $w_{i}$. See details in Section~\ref{tree_gru}.
\end{itemize}

As we can see, SDP is a path-based syntax representation, while TPF and Tree-GRU are tree-based representations that aim to capture global information of a parse tree. To incorporate syntactic information, we accumulate the encoding of dependency representations with the basic input $\bm{q}$ of each question. The architecture of our system is demonstrated in Figure~\ref{fig:querygraph}.

\subsection{Path-Based Representation: SDP}
\label{sdp_method}
\begin{figure*}
  \centering
  \includegraphics[width=15cm]{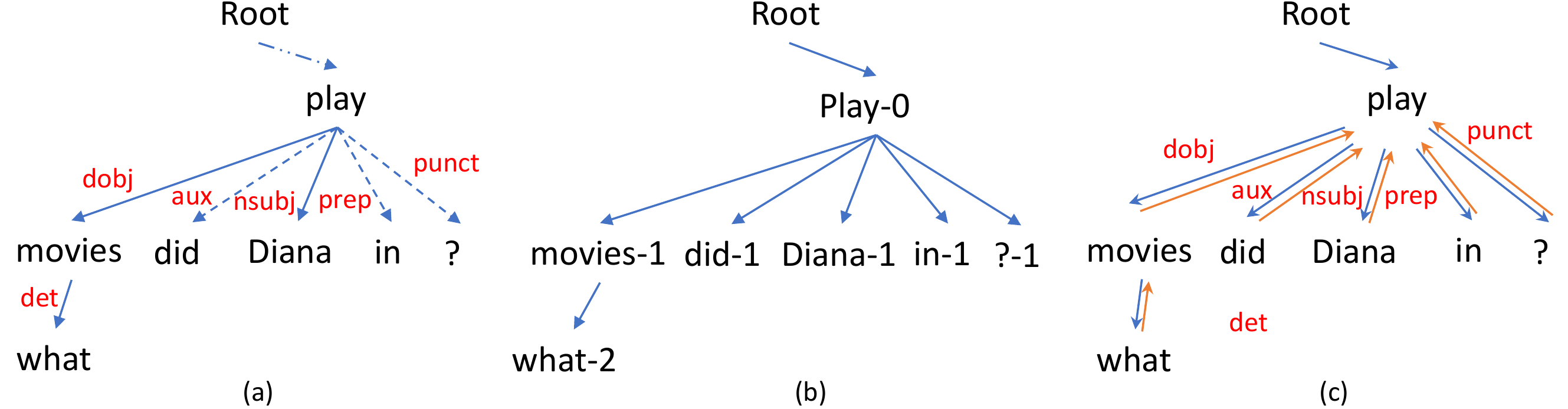}
  \caption{Illustration of the encoding methods of SDP (a), TPF (b), and Tree-GRU (c).}
  \label{fig:sdp_dpl_tgru_tree}
\end{figure*}

A shortest dependency path describes the path information of two specified words in a dependency  tree. 
Xu et al.~\shortcite{Xu2015ClassifyingRV} for the first time utilize neural network models to encode the shortest dependency path information between entities for the task of relation classification. For the task of KBQA, we can take the same approach for representing the dependencies between the focus words and the answer word, as illustrated in Figure~\ref{fig:sdp_dpl_tgru_tree} (a). The answer word usually starts with ``wh-" in a question, such as ``what" and ``who". The advantage of using shortest dependency paths as syntax representation is that it can capture salient information of the focus words to avoid interfering from non-important words. By using the shortest path method, the questions with multiple entities can be decomposed into several shortest dependent paths each of which corresponds to a focus word. 

Luo et al.~\shortcite{Luo2018KnowledgeBQ} apply the SDP method onto the KBQA task and have achieved remarkable improvements. Here, we adopt the same strategy. Specifically, given a question, we first recognize four types of focus words which are named entities, type words, time words, and ordinal words. We then collect a shortest dependency path for each focus word. Figure~\ref{fig:sdp_dpl_tgru_tree}(a) presents an illustrating example of a specific entity word, where the shortest dependency path corresponding to the word ``Diana" is ``what $\vec{det}$ movies $\vec{dobj}$ play $\vec{nsubj}$ Diana". To encode such a dependency path, we use pre-trained models like GloVe to initialize tokens in the path and then apply BiGRU to obtain the vector representation of the path. For focus words of other types, a similar process can be applied. For multiple different path representations, the maxpooling operation is used to extract the final vector. We finally get the encoding of the question using SDP, denoted as $\bm{x}_1^{syn}$.

\subsection{Tree-Based Representation: TPF}
\label{tpf_method}
Yang et al.~\shortcite{Yang2016APE} propose the tree position feature (TPF) for the task of relation classification, which could express the position information of each word in a dependency tree. Specifically, given a word $w_i$, we regard the root of a parse tree as a reference node and define the length of the path that connects the reference node and $w_i$ as the corresponding position feature. As shown in Figure \ref{fig:sdp_dpl_tgru_tree}(b), ``play" is the reference node, and the position feature of the word ``what" is $2$. Then, we convert the position features into  vectors by using random initialization.

In this way, for each word $w_i$ in a question, we can get its word embedding $\bm{w}_i$ and the corresponding tree position vector $\bm{d}_i$. Then we concatenate $\bm{w}_{i}$ and ${\bm{d}}_i$ to get the expression $\bm{z}_i=({{\bm{w}}_{i}},{\bm{d}}_i)$ for $w_i$. We apply BiGRU to encode the whole question and get the TPF-based representation, denoted as $\bm{x}_2^{syn}$.

\subsection{Tree-Based Representation: Tree-GRU}
\label{tree_gru}
The approach to Tree-GRU based representation consists of two steps. The first step is pre-training on a large set of auto-parsed questions which learns the embeddings of dependency edges. The second step is the application of Tree-GRU to learn vector representations of the words in a question.

\subsubsection{Pre-training of Edge Embeddings} 
We collect a set of 4.99M questions from multiple data sources, including SimpleQuestions ~\footnote{https://github.com/davidgolub/SimpleQA/tree/master/datasets/}, ComplexQuestions ~\footnote{https://github.com/JunweiBao/MulCQA/tree/}, Quora ~\footnote{https://www.kaggle.com/c/quora-insincere-questions-classification/data}$^,$~\footnote{https://www.kaggle.com/quora/question-pairs-dataset}, LC-QuAD 1.0~\footnote{http://lc-quad.sda.tech/lcquad1.0.html}, and LC-QuAD 2.0~\footnote{http://lc-quad.sda.tech/}. We then utilize the Stanford Parser to parse each question. 
With the resulting set of auto-parsed questions, the skip-gram neural network language model~\cite{Mikolov2013EfficientEO} is used to pre-train embeddings of dependency edges. For an edge $a \xrightarrow{\textit{e}} b$ in a dependency tree, $a$ and $b$ are the head and tail of the edge $e$ respectively. Then we get the neighboring edges of the edge $a \xrightarrow{\textit{e}} b$ according to the dependency tree: 
$$\{ e_a\} \cup \{ e_b^1,e_b^2,\ldots, e_b^i,\ldots, e_b^n\},$$
where $\{ e_a\}$ represents the set of incoming edges of the node $a$ (except for the root node, each tree node has only one incoming edge), $\{e_b^i\}$ represents the set of outgoing edges of the node $b$, and $n$ refers to the number of outgoing edges of node $b$. The learning objective of skip-gram on the training data is to use the edge $a \xrightarrow{\textit{e}} b$ to predict neighboring edges . We finally obtain the pre-trained edge embeddings $\bm{E}_{e} \in \mathbb{R}^{|\bm{V}_e| \times d}$, where $|\bm{V}_e|$ denotes the number of all unique edges in auto-generated dependency trees and $d$ represents the dimension size of the pre-trained embeddings.

\subsubsection{The Tree-GRU encoding}
Tree-GRU, as a variant of Tree-RNN, is an extension from sequence-structured GRU to tree-structured GRU~\cite{chen_etal_acl_2017}. Tree-GRU can learn global vector representations for each word in a question in a top-down and bottom-up learning process, as shown in Figure~\ref{fig:sdp_dpl_tgru_tree} (c). In the bottom-up learning process, the computation of $\bm{h}_{i}^\uparrow$ of each word $w_i$ is based on its child nodes. The detailed equations are as follows. 

\begin{align}
&\bar{\bm{h}}_{i,L}^\uparrow = \sum_{j \in lchild(i)} \bm{h}_j^\uparrow ,  &\bar{\bm{h}}_{i,R}^\uparrow = \sum_{k \in rchild(i)} \bm{h}_k^\uparrow
\end{align}

\begin{equation*}
\begin{aligned}
& \bm{r}_{i,L}  =\sigma(\bm{W}^{rL}\bm{l}_{i}+\bm{U}^{rL} 
\bar{\bm{h}}_{i,L}^\uparrow +\bm{V}^{rL} 
\bar{\bm{h}}_{i,R}^\uparrow )\textrm{,} & \\
& \bm{r}_{i,R}  =\sigma(\bm{W}^{rR}\bm{l}_{i}+\bm{U}^{rR} 
\bar{\bm{h}}_{i,L}^\uparrow +\bm{V}^{rR} 
\bar{\bm{h}}_{i,R}^\uparrow )\textrm{,} & \\
& \bm{z}_{i,L}  =\sigma(\bm{W}^{zL}\bm{l}_{i}+\bm{U}^{zL} 
\bar{\bm{h}}_{i,L}^\uparrow +\bm{V}^{zL} 
\bar{\bm{h}}_{i,R}^\uparrow )\textrm{,} & \\
& \bm{z}_{i,R}  =\sigma(\bm{W}^{zR}\bm{l}_{i}+\bm{U}^{zR} 
\bar{\bm{h}}_{i,L}^\uparrow +\bm{V}^{zR} 
\bar{\bm{h}}_{i,R}^\uparrow )\textrm{,} & \\
& \bm{z}_{i} =\sigma(\bm{W}^{z}\bm{l}_{i}+\bm{U}^{z} 
\bar{\bm{h}}_{i,L}^\uparrow +\bm{V}^{z} 
\bar{\bm{h}}_{i,R}^\uparrow )\textrm{,} & \\
& \hat{\bm{h}}_{i}^\uparrow =\tanh(\bm{W}\bm{l}_{i}+\bm{U}(\bm{r}_{i,L} \odot \bar{\bm{h}}_{i,L}^\uparrow) + \bm{V}(\bm{r}_{i,R} \odot \bar{\bm{h}}_{i,R}^\uparrow))\textrm{,} & \\
& \bm{h}_{i}^\uparrow = \bm{z}_{i,L} \odot \bar{\bm{h}}_{i,L}^\uparrow+ \bm{z}_{i,R} \odot \bar{\bm{h}}_{i,R}^\uparrow + \bm{z}_{i} \odot \hat{\bm{h}}_{i}^\uparrow \textrm{,}&
\end{aligned}
\end{equation*}
where $lchild (i)$ and $rchild (i)$ refer to the set of left and right children of node $w_i$, $\bm{l}_{i}$ represents the dependency edge label between $w_i$ and its parent node, $\bm{W}$, $\bm{U}$ and $\bm{V}$ are trainable parameters of the model. 
In the same way, we can get the encoding vector $\bm{h}_i^\downarrow$ of each word in the question through the top-down encoding method. Finally, we concatenate the resulting hidden vectors obtained by these two encoding methods to get the final representation of a word:

\begin{align}
    \bm{t}_{i} &= \bm{h}_i^\uparrow \oplus \bm{h}_i^\downarrow
\end{align}
By concatenating the word embedding $\bm{w}_i$ and Tree-GRU embedding $\bm{t}_{i}$, we cat get the vector representation  $\bm{z}_i=\{\bm{w}_i,\bm{t}_{i}\}$ for the word $w_i$. We then apply BiGRU to obtain the Tree-GRU based representation of the question, denoted as $\bm{x}_3^{syn}$.

\subsection{Syntax-Aware System}
For a question, we have basic representation $\bm{q}$ (described in Section \ref{sec:basicencode}) and three types of syntax-based representations, $\bm{x}_1^{syn}$ (SDP), $\bm{x}_2^{syn}$ (TPF), and $\bm{x}_3^{syn}$ (Tree-GRU). Then, we directly accumulate these representations to obtain new representation of the question. Finally, we adopt the same training objective and optimization algorithm with the baseline.

\section{Experiments}
In this section, we evaluate the effect of different syntax-based representations for KBQA and conduct the experiments on a widely used dataset.

\subsection{Experimental Setup}
\subsubsection{Datasets}
We adopt the benchmark dataset WebQuestions~\cite{Berant2013SemanticPO}, which has been widely used in many recent works for the task of KBQA~\cite{Qu2018QuestionAO,Luo2018KnowledgeBQ}. The WebQuestions dataset contains both simple and complex questions, which are collected from Google Suggest API. We use the standard data split: 3,778 question-answer pairs for training and 2,032 pairs for testing. To tune the parameters of the systems, we randomly select about 20\% (755 pairs) from the training set as the development test. 

As for the knowledge base, we use the complete Freebase dump~\footnote{http://dwz.win/qRM or https://github.com/syxu828/QuestionAnsweringOverFB/} as the settings in~\newcite{Luo2018KnowledgeBQ}. The knowledge base contains 46M entities and 5,323 predicates, and we use Virtuoso-engine~\footnote{http://virtuoso.openlinksw.com/} to host the knowledge base. To parse questions, we use Stanford Parser V3.5.1~\footnote{https://nlp.stanford.edu/software/lex-parser.html}. To initialize word embeddings, we utilize GloVe word vectors~\cite{Pennington2014GloveGV} and word vectors from ELMo \cite{Peters2018DeepCW}.
\subsubsection{Hyperparameters}
We tune the hyperparameters of the baseline and our systems on the development set. We adopt the brute-force grid search to decide optimal hyperparameters. Specifically, the dimension of GloVe word vectors is set to 300 and the dimension of ELMo is set to 1,024. The Adam optimizer adopts a batch size of 32. The dropout rate is set to 0.1 when using the TPF method and the Tree-GRU method. We also use a dropout layer after the embedding layer.  In order to render the results reliable, we run each experiment three times by using different random seeds and report the average value as the final result.

\begin{table*}[tb]
\begin{minipage}[b]{77mm}
\begin{tabular}{p{5cm} l}
\hline
Question Representation & F1($\%$) \\
\hline
Baseline (QUE) & 52.33 \\

QUE+SDP   & 52.96 \\
QUE+TPF   & 52.58 \\
QUE+Tree-GRU   & 52.12 \\
QUE+SDP+TPF   & 53.27 \\
QUE+SDP+Tree-GRU   & 53.35 \\
\hline
Baseline (QUE)+ELMo & 52.83 \\
QUE+SDP+TPF+ELMo  & 53.59 \\
QUE+SDP+Tree-GRU+ELMo  & 53.63 \\
\hline
\end{tabular}
\caption{Main results.}
\label{tab:dep_baseline}
\end{minipage}
\begin{minipage}[b]{77mm}
\begin{tabular}{p{5cm} c}
\hline
Method & F1($\%$) \\
\hline
\newcite{Dong2015QuestionAO} & 40.8 \\
\newcite{yaoetal:2015} & 44.3 \\
\newcite{bastetal:2015} & 49.4 \\
\newcite{berantetal:2015} & 49.7 \\
\newcite{Yih2015SemanticPV} & 52.5 \\
\newcite{Bao2016ConstraintBasedQA} & 53.3 \\
\newcite{jainetal:2016} & 55.6 \\
\newcite{cuietal:2017} & 34.0 \\
\newcite{Luo2018KnowledgeBQ} & 52.7 \\
\newcite{chenetal:2019} & 51.8 \\
\midrule
Baseline (QUE) & 52.33 \\
QUE+SDP+Tree-GRU  & 53.35 \\
QUE+SDP+Tree-GRU+ELMo  & 53.63 \\
\hline
\end{tabular}
\caption{Comparison with previous works.}
\label{tab:results}
\end{minipage}
\end{table*}

\subsection{Main Results}
Here, we demonstrate the effect of syntax-based  representations. We compare the following systems in the experiments: (1) Baseline (QUE): in the baseline system, we use the word vectors of GloVe to initialize word embeddings of questions. (2) QUE+SDP: we add the information of short dependency path (SDP) to Baseline. (3) QUE+TPF: we add the information of tree position features (TPF) to Baseline. (4) QUE+Tree-GRU: we add the information provided by Tree-GRU to Baseline. (5) QUE+SDP+TPF: we add the information of tree position features (TPF) to QUE+SDP. (6) QUE+SDP+Tree-GRU: we add the information provided by Tree-GRU to QUE+SDP.



Table \ref{tab:dep_baseline} shows the results. From the table we can find that adding SDP and TPF solely yields an improvement of $0.63\%$ and $0.25\%$ F1 scores respectively, while adding Tree-GRU slightly underperforms than Baseline (QUE). Then, we further use the syntax-based representations together and show better performance. Finally, our best system (QUE+SDP+Tree-GRU) outperforms Baseline by an absolute improvement of $1.02\%$ F1 score. These facts indicate that the syntax-based representations are effective in improving the KBQA systems.


To further improve our systems, we utilize vectors of ELMo for word representations of questions. The results are depicted in Table~\ref{tab:dep_baseline}. From the results we can see that using ELMo results in a stronger baseline with an improvement of $0.6\%$ F1 score. In addition, ELMo can further enhance our previous best system (QUE+SDP+Tree-GRU) by an improvement of $0.28\%$ F1 score. Our new best system (QUE+SDP+Tree-GRU+ELMo) achieves the final score of $53.63\%$ with an improvement of $1.3\%$ over Baseline. These facts indicate that our proposed syntax-based representations are complement with the pre-trained language models.
  

\subsection{Comparison with Previous Approaches}
Table~\ref{tab:results} shows the comparison of our systems with previous works. The results show that our Baseline is a strong system and our final system achieves the second best score in the table. In the work of~\newcite{jainetal:2016}, they use the relatively complex fact memory network which has better reasoning ability for complex questions. In future, we may use our proposed syntax-based representations in their system.

\subsection{Analysis and Discussions}

\begin{figure*}
    \centering
    \includegraphics[width=15.5cm]{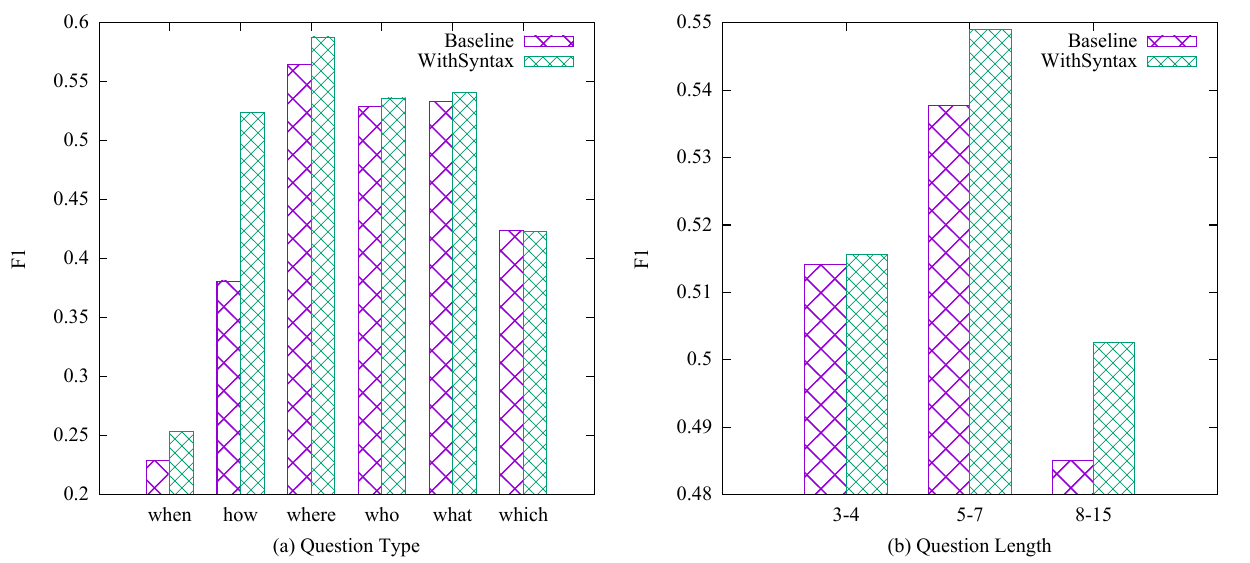}
    \caption{Performance comparison of baseline and syntax-aware system.}
    \label{fig:analysis}
\end{figure*}

We further compare the performance between Baseline and our system (QUE+SDP+Tree-GRU) to analyze the effect of syntactic information on question types and question lengths.

The results on question types are shown in Figure \ref{fig:analysis}(a). From the figure, we find that the syntax-aware system provides better performance on five types, while two systems yield similar scores for type ``\emph{which}". This indicates that the syntax-based representations can provide useful information for the general cases. 



The results on question lengths are shown in Figure \ref{fig:analysis}(b). we split the test set into three sets by the question length: SHORT(3-4), MID(5-7), and LONG(8+). From the figure,
we find that the improvement gap is larger on the LONG set than that on the MID and SHORT sets. The reason might be that the dependency trees can provide useful connections for the words with long distance.

\begin{figure*}
    \centering
    \includegraphics[width=8cm]{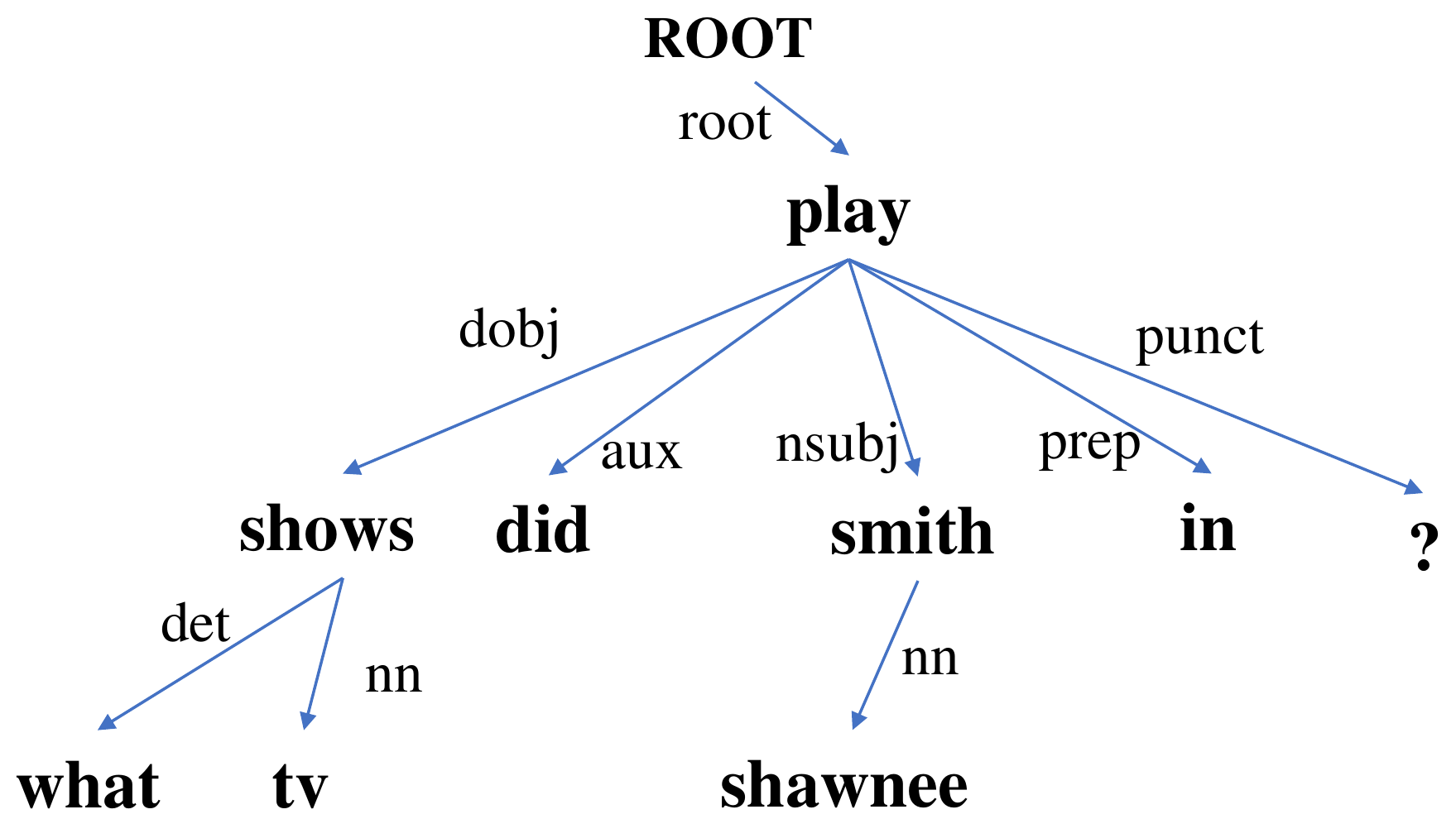}
    \caption{Case study: dependency tree of a question.}
    \label{case_tree}
\end{figure*}

\subsection{Case Study}
In this section, we check the results in more detail. We find that the syntax information is helpful for the cases where the constraint words are far away from the predicate words. Figure \ref{case_tree} shows the dependency tree of a real example for case study, where the question is ``What TV shows did Shawnee Smith play in?", in which ``TV shows" is the constraint words and ``play" is the predicate word. The baseline system can not identify the connection between ``TV shows" and ``play", while the syntax-aware system can have this information as shown in Figure \ref{case_tree}. Due to the lack of sufficient information, the baseline system  returns the wrong relationships ``film.actor.film" and ``film.performance.film", and fails to predict the correct answer. In contrast, the syntax-aware system can map to the correct relationships ``tv.tv\_actor.starring\_roles" and ``tv.regular\_tv\_appearance.series", and obtain the correct answer.

\section{Related Work}

In the field of KBQA, there are two group of representative approaches: systems based on query graph generation and systems based on semantic parsing. The main idea of the former group is to apply information extraction techniques to construct structured queries which can be issued to knowledge bases to get the answer. A bunch of previous works have been conducted in this direction, includeing~\newcite{yaoetal:2014},~\newcite{Bordes2015LargescaleSQ},~\newcite{Dong2015QuestionAO},~\newcite{Bao2016ConstraintBasedQA},~\newcite{Xu2016QuestionAO},~\newcite{Hu2018AnsweringNL}, and~\newcite{Luo2018KnowledgeBQ}. The main difference between the approaches is the representation of query graphs. In contrast, the latter group of systems relies on semantic parsing techniques which often suffer from the lack of training data for build semantic parsers. The output of semantic parsing are queries in the form of formal meaning representation.~\newcite{Berant2013SemanticPO} and \newcite{Kwiatkowski2013ScalingSP} have proposed different models for semantic parsing of natural language questions. Our work focuses on incorporating rich syntax into a system that is built on query graph generation.

\section{Conclusion}
In this paper, we show that utilizing the syntax-based representations can have a substantial effect on the performance of KBQA. In particular, we demonstrate that our systems consistently outperform a strong baseline system using ELMo. In our approach, we consider two types of 
syntax-based representations, path-based and tree-based representations. As for the tree-based representations, we model the whole syntactic trees to obtain global information. Furthermore, the experimental results show that two types of representations are complementary to each other in our final system.



\bibliographystyle{coling}

\end{document}